%% file: paper.tex
\relax
\documentclass[letterpaper]{article} 
\usepackage{aaai21}  
\usepackage{times}  
\usepackage{helvet} 
\usepackage{courier}  
\usepackage[hyphens]{url}  
\usepackage{graphicx} 
\usepackage{natbib}  
\usepackage{caption} 
\usepackage{algorithm}
\usepackage{algpseudocode}
\usepackage{algorithmicx}
\usepackage{tabularx}
\usepackage{amsmath}
\usepackage{enumitem}
\usepackage{graphicx}
\usepackage{bbding}
\usepackage{fontawesome}
\usepackage{eqparbox}
\usepackage{multirow}
\usepackage{hhline}
\usepackage{rotating}
\usepackage{soul}
\algtext*{EndWhile}
\algtext*{EndIf}
\algtext*{EndFor}

\let\oldReturn\Return
\renewcommand{\Return}{\State\oldReturn}

\algnewcommand\algorithmicto{\textbf{to}}
\algdef{SE}[FOR]{ForTo}{EndFor}[2]{%
  \algorithmicfor\ #1\ \algorithmicto\ #2\ \algorithmicdo
}{%
  \algorithmicend\ \algorithmicfor
}
\algrenewcommand\algorithmicdo{}

\newcommand{\blue}[1]{\textcolor{blue}{#1}}
\newcommand{\red}[1]{\textcolor{red}{#1}}

\usepackage{color, soul}

\soulregister\cite7
\soulregister\ref7
\soulregister\pageref7

\setstcolor{red}

\title{GraphPlan: Story Generation by Planning with Event Graph}

\author{
    Hong Chen\textsuperscript{\rm 1,3}, Raphael Shu\textsuperscript{\rm 1}, Hiroya Takamura\textsuperscript{\rm 2,3}, Hideki Nakayama \textsuperscript{\rm 1}
    \\
}
\affiliations{
    \textsuperscript{\rm 1}The University of Tokyo, 
    \textsuperscript{\rm 2}Tokyo Institute of Technology,\\
    \textsuperscript{\rm 3}National Institute of Advanced Industrial Science and Technology, Japan\\ 
    \{chen, nakayama\}@nlab.ci.i.u-tokyo.ac.jp,       raphael@uaca.com,     takamura.hiroya@aist.go.jp\\

}
\begin{document}

\maketitle

\input{abstract}

\input{introduction}

\input{related}

\input{background}

\input{method}

\input{experiment}

\input{discussion}

\input{conclusion}

\bibliography{cite}
\end{document}

%% file: abstract.tex
\begin{abstract}

Story generation is a task that aims to automatically produce multiple sentences to make up a meaningful story. This task is challenging because it requires high-level understanding of semantic meaning of sentences and causality of story events. Naive sequence-to-sequence models generally fail to acquire such knowledge, as the logical correctness can hardly be guaranteed in a text generation model without the strategic planning. In this paper, we focus on planning a sequence of events assisted by event graphs, and use the events to guide the generator. 
Instead of using a sequence-to-sequence model to output a storyline as in some existing works, we propose to generate an event sequence by walking on an event graph. The event graphs are built automatically based on the corpus. 
To evaluate the proposed approach, we conduct human evaluation both on event planning and story generation.
Based on large-scale human annotation results, our proposed approach is shown to produce more logically correct event sequences and stories.

\end{abstract}

%% file: introduction.tex
\section{1\quad Introduction}

Narrative Intelligence \cite{NI} is one form of Humanistic Artificial Intelligence that requires the system to organize, comprehend, and reason about narratives and produce meaningful responses.
Story generation tasks can be considered as a test bed for examining whether a system develops a good understanding of the narratives.

Other than leaving the model to output random sentences, the model is usually given a specific topic (e.g., title or prompt) or visual information (e.g., image or video).
One straight-forward approach for these story generation tasks is to leverage a sequence-to-sequence model to predict sentences sequentially. Although the model can be trained to capture the word-prediction distribution from training data, it has two serious drawbacks when applied to generate stories:
1) Conditional Language model (i.e., decoder) tends to assign high probabilities to generic, repetitive words, especially when beam search is applied in the decoding phase~\cite{holtzman2019curious};
2) Sequence-to-sequence models often fail to produce logically correct stories.

\begin{figure}[!t]
\centering
\includegraphics[scale=0.18]{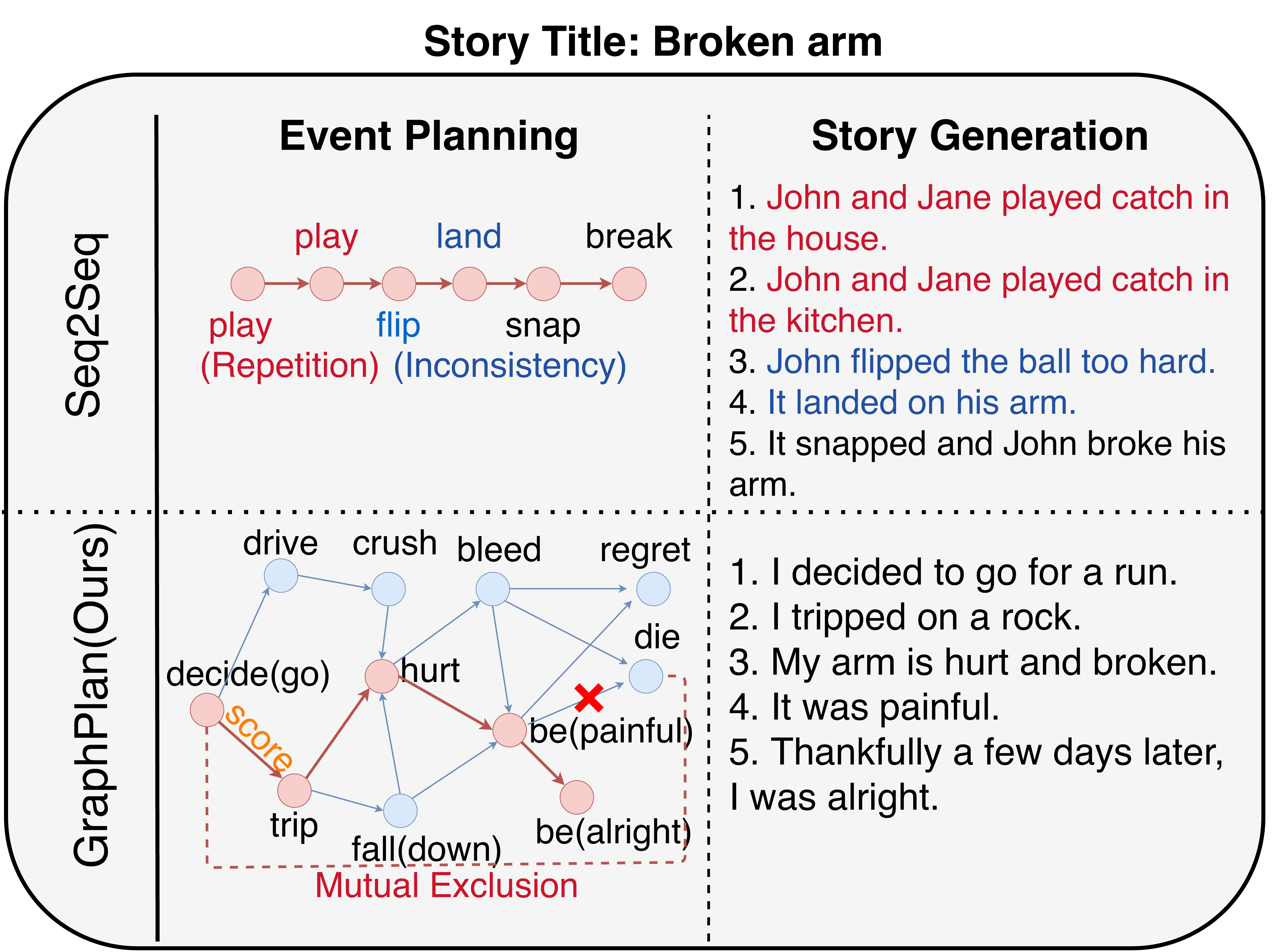}
\caption{
Comparison between sequence-to-sequence model and GraphPlan (ours). Two problems in the sequence-to-sequence model when generating events: \red{Repetition} and \blue{Logical Inconsistency}. Repeated words (e.g., play) in the storyline result in the repeated sentences in the generated stories. Besides, The logic between ``land'' and ``snap'' lacks causality, thus generating incoherent stories. On the contrary, our GraphPlan method does not rely on any language model, it applies beam search on the event graph based on a well-designed score function. The mutually exclusive set further ensures the global logical consistency for the planned events.}
\label{fig:problems}
\end{figure}

Recently, huge interests have been aroused to decompose story generation into two phases: planning and generation~\cite{plan_write,plan_write_revise,xu2018skeleton,fan2019strategies}. 
Planning~\cite{plan,riedl2010narrative} creates a high-level abstraction or a blueprint to encourage the generator to focus on the flow of a story, similar to making an outline before writing. 
The planned elements are referred to as \textit{events} in many papers. However, the detailed definition of events varies. 
For instance, an event can be represented as a verb argument pair (e.g., {\it (admits, subj)})~\cite{chambers2008unsupervised}, a tuple of subject, verb, object and modifier or ``wildcard'' (e.g., {\it(PERSON0, correspond-36.1, empty, PERSON1)})~\cite{event,event2} or a reconstructed verb phrase (e.g., {\it decide(go)})~\cite{two_discourse}.
In this paper, we follow \citet{two_discourse} to represent an event with verb phrases.

Previous works~\cite{plan_write} have shown that if the events are well-planned, then the correctness of generated stories is almost guaranteed, and furthermore, the stories can be easily controlled by modifying the events.
However, existing approaches~\cite{plan,plan_write_revise,event,event2} regard event generation as an abstracted version of story generation. In other words, they treat each event as one token and use a sequence-to-sequence model to make a plan of the events. 
Our preliminary experiments show that repetition and logical inconsistency problems happen in the event sequence and same problems occur in the generated stories.
Figure~\ref{fig:problems} shows an example using sequence-to-sequence in event planning. We can see that the both events and generated stories are repeated and illogical.


In this paper, instead of leveraging a sequence-to-sequence model on event planning, we propose a planning method \textbf{GraphPlan}. To plan the events, GraphPlan walks on a topic-specific event graph with beam search. 
Event graphs are adopted for story generation even before the emergence of neural-based models \cite{plotgraph2,plotgraph3,plotgraph4,plotgraph5,plotgraph}.
An event graph represents the logical flow of events based on the facts presented in a corpus. With a learned event graph, we can walk on it and produce a reasonable event sequence. 
We follow the graph setting in \citet{plotgraph}, in which each graph is composed of event nodes, connections and a set of mutually exclusive events. 

\begin{figure}[!t]
\centering
\includegraphics[scale=0.18]{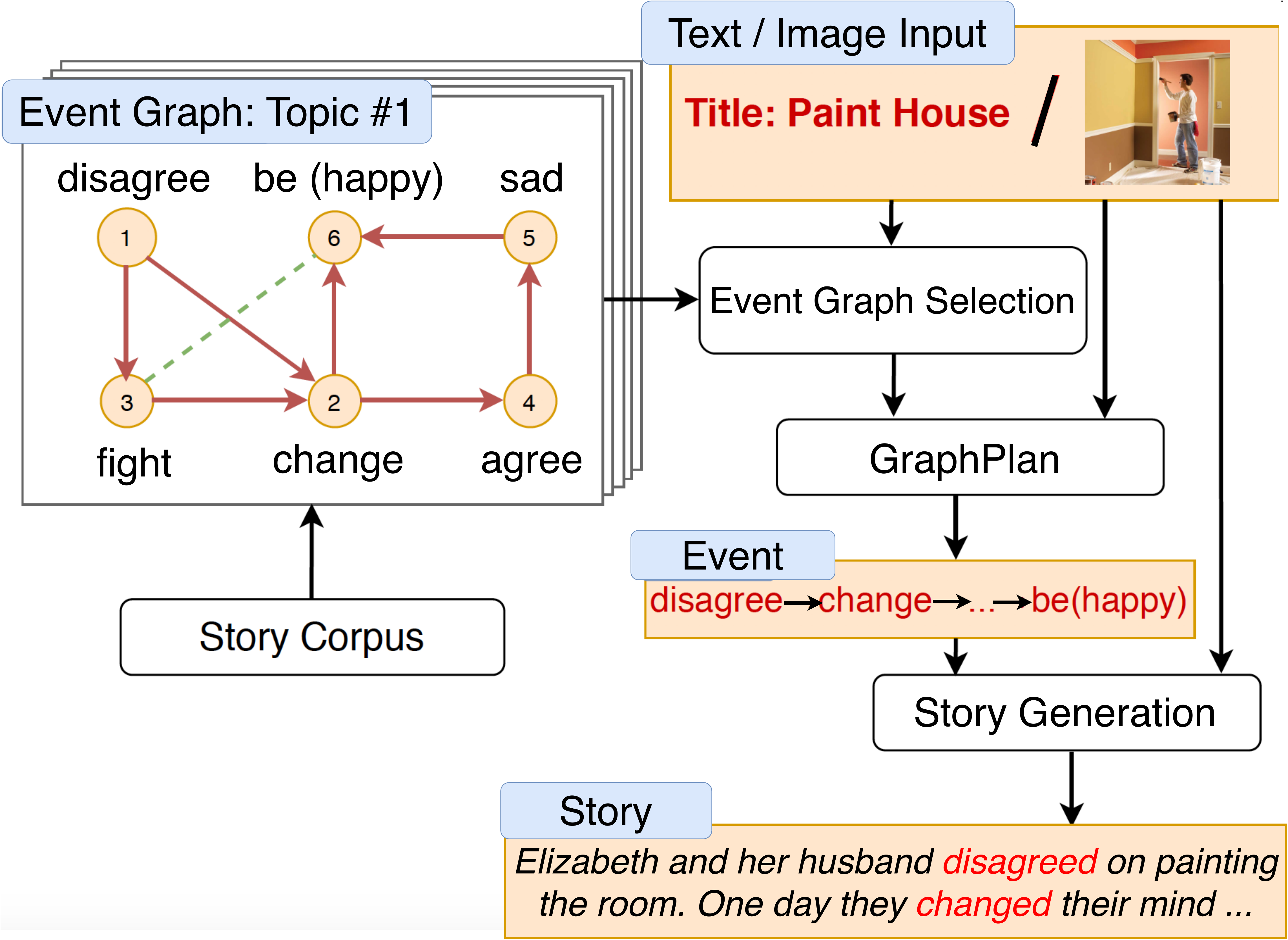}
\caption{Overview of our approach. In the preprocessing step, we cluster the stories into $K$ topics and build an event graph for each topic. In the planning step, a event graph selection module select a event graph based on the input. Then a related event graph is retrieved. The event planning model generates a sequence of events. Finally, based on the input and the planned events, a story generation module generates the story. The dash line denotes the mutually exclusive events that can hardly coexist in one storyline.}
\label{fig:system}
\end{figure}

To generate a story, we first identify the topic based on the input (e.g., title or image) and retrieve a related event graph. We then plan the events by running beam search with a score function that takes the event-event coherence and input-event coherence into account. 
Finally, a \textit{story generation} module transforms the planned event sequence into a readable story.
Figure~\ref{fig:problems} shows an example of using GraphPlan and Figure~\ref{fig:system} depicts the whole pipeline of our proposed approach.

We conduct experiments on open story generation to evaluate how event graphs benefit the task. Our approach is shown to significantly outperform baseline models that generate events with sequence-to-sequence models in terms of logical consistency. We also conduct Story Cloze Test to further validate the effectiveness of the event graphs and the mutually exclusive sets.
Our contributions can be summarized as follows:

\begin{itemize}[noitemsep,topsep=0pt]
    \item We propose a score-based beam search approach to plan story events with an event graph.
    \item Comparing to baseline models, our graph-based planning approach results in much better logical correctness in story generation tasks according to the human evaluation.
    \item Experiments on Story Cloze Test directly confirms the high accuracy of the proposed event planning approach.
\end{itemize}

%% file: related.tex
\section{2\quad Related Work}

\paragraph{Planning for Story Generation}
Several approaches have been explored to plan a skeleton of the story before actual generation. Before the emergence of neural-based models, \citet{plan5} and \citet{plan3} attempted to use hand crafted rules to arrange actions into character sequences. Recently, with the help of neural sequence-to-sequence models, \citet{xu2018skeleton} proposed to generate multiple key phrases and expand them into a complete story. A built-in key phrases generation module is used in their model architecture.
In contrast to \citet{xu2018skeleton}, some works explicitly plan a sequence of events \cite{event, event2, controllable}, keywords \cite{plan_write, ippolito-etal-2019-unsupervised} or actions \cite{fan2019strategies} before generating the story based on the planned items.

All of these planning models rely on a language model for planning without following an external structure of events, which resulted in degraded performance \cite{holtzman2019curious}. Compared with these works, the main contribution of this paper is to propose a planning method based on automatically created event graphs. Instead of a language model, we use score-based beam search to generate a sequence of events by walking on the graph.



\paragraph{Graph-based Story Planning}
Event graph is a variant of plot graph whose nodes represent events. A quantity of research made progress on generating stories from plot graphs~\cite{plotgraph2,plotgraph3,plotgraph4,plotgraph5,li-etal-2019-coherent}.
\citet{plotgraph} proposed a plot graph on story generation tasks which is most related to our work. They crowd-sourced the story corpus and manually created plot nodes and edges in the graph. In their graph, mutually exclusive events are not allowed to be present in the same story. 

In this work, both the event graphs and mutually exclusive sets are automatically generated. We further propose an event planning method taking into account the relations between events and various inputs (i.e., title or image).

%% file: background.tex
\section{3\quad Event Graph Construction}
As a preprocessing step, we first extract events automatically from a corpus. Then, we divide the corpus into several topics. Finally, we build an event graph for each topic.

\paragraph{Data-based Event Extraction} Following~\citet{two_discourse}, we represent each event with a verb phrase.
Unlike other representations, a verb phrase is the minimum unit in one sentence which is abstract, simple and comprehensible for humans. 
From our observation, this representation does not have a severe sparseness problem. In statistics, each event can connect over 3 next possible events on average. Please note that our work does not investigate event representation, we focus on planning a more logical event sequence.
Specifically, as a preprocessing step, we parse all sentences with semantic role labeling and extract the verb phrases. 
If an extracted verb has an argument with semantic role ``AM-NEG'' (negation) for a verb, we add {\it (not)} before it (e.g., {\it (not)take}). If a verb is followed by a preposition, we append the prepositional word to the verb (e.g., {\it take(over)}). If the label is ``AM-PRD’’ (secondary predicate), we make an event from it (e.g., {\it be(excite)}). Finally, if two verbs are close to each other within five-word distance in the corpus, we combine them to make an event (e.g., {\it decide(buy)}). All words in the event are stemmed with NLTK \cite{nltk}. 

\paragraph{Topic Modelling}
Generally, a story dataset contains a variety of topics ranging from animal, health, to robbery. 
Here, we use Latent Dirichlet Allocation (LDA)~\cite{lda} to infer the topics in the corpus.
Considering that the relation between events drastically changes according to the topic, in this work, we build an independent event graph for each topic. 
Formally, we denote $e_1^k,\ldots,e_{t}^k$ the event set from the stories that belong to the k-th topic $\mathcal{T}^k$ in the corpus. These events would be used as nodes for the event graph of $\mathcal{T}^k$. 
LDA clusters the stories and thus reduces the amount of unique events in each graph, which will make the graph walking algorithm more efficient.

\paragraph{Event Connection}
After collecting the events from a corpus for each topic, we need to find connections among these events to build a graph. The connections are represented as directed edges whose direction indicates possible next events. 
In practice, if events $e_i$ and $e_j$ occur adjacently in the text, we add an edge $e_i \rightarrow e_j$.
An example of an event graph can be found in Figure~\ref{fig:system}.

\paragraph{Mutually Exclusive Set}
Following the graph setting in \citet{plotgraph}, there are events (e.g. ``die'' and ``be(happy)'') that are mutually exclusive and cannot be placed in one story. These mutually exclusive relations are considered as exceptions and difficult to be represented along with the event graph. We create a held-out set consisting of mutually exclusive event pairs for each graph.

\begin{figure}[t]
\centering
\includegraphics[scale=0.12]{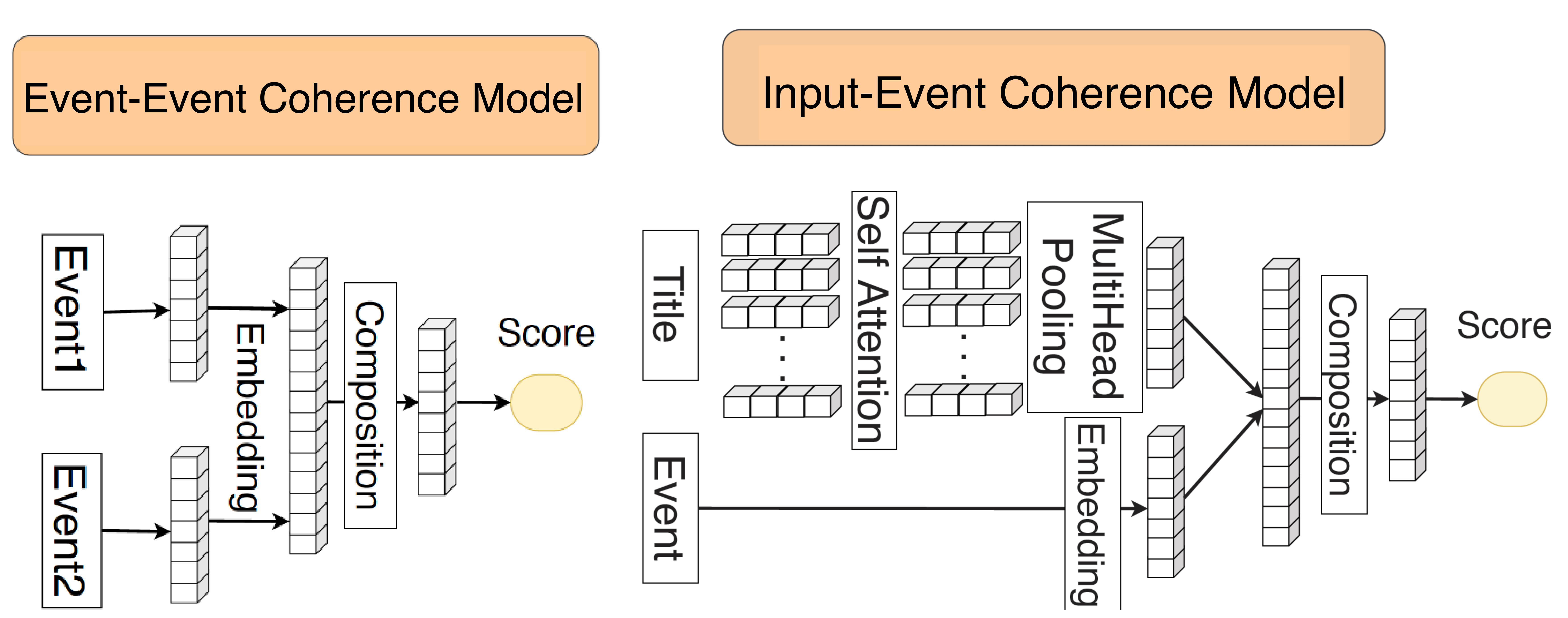}
\caption{Coherence models we used in this paper. The event-event coherence model outputs a coherence score for two events. The input-event coherence model takes a title and an event as input. Both coherence models finally produce a score within 0 to 1. These coherence scores decide the next event when running beam search.}
\label{fig:coherence}
\end{figure}

To identify these mutually exclusive events from the constructed graphs, we prepare an event-event coherence model to detect the coherence score between two events. We prevent two events with low coherence scores from coexisting in the planned events. 
The model architecture is based on compositional neural networks~\cite{scriptlearning}, as shown in Figure~\ref{fig:coherence}.
The model takes two events ($e_i, e_j$) represented with unique embeddings and outputs a coherence score normalized with the sigmoid function $f_{event}(e_i, e_j) \in [0,1]$. 
We use contrastive training to optimize the model. Here, positive examples are the events extracted from the same story or title, whereas negative examples are randomly sampled from the events in different stories. 
Let $(e_i, e_j)$ denote a positive pair of events and $\tilde{e}_j$ denote a randomly sampled event. The training loss for the event-event coherence model is defined as:
\begin{align}
    L_{event} =  \max(0, - f_{event}(e_1, e_2) + f_{event}(e_1, \tilde{e}_2) + m)
\end{align}
where $m$ is a fixed margin. Finally, we consider two events are mutually exclusive if the coherence score falls below a certain threshold $\tau$.

On average, after taking the mutually exclusive sets into account, each event graph can still plan over one million different possible event sequences. Please refer to the supplementary materials for more statistics details of event graphs. Additionally, these in-topic event graphs can be hierarchically combined into a larger graph if the model is required to generate longer discourse-level stories. This will be a future direction of our work.

%% file: method.tex
\section{4\quad GraphPlan: Planned Story Generation with Event Graph}
\label{sec:graphplan}
In this section, we describe our approach for planned story generation. 
We separated the whole pipeline into two steps:
1) Our GraphPlan walks on the event graph and produces a sequence of events as a blueprint of the story. 
2) The story generation module then finalizes the texts following the planned events.




\subsection{4.1\quad GraphPlan}
Till now, each topic has a corresponding event graph. Before story generation, we propose GraphPlan to plan event sequences from the event graph.
These planned events will be used to guide the story generation module in the next step. 
GraphPlan contains two steps. 1) Selecting an event graph for the input (i.e. title or image). 2) Generating an event sequence by walking on the graph.
\paragraph{Event Graph Selection} Firstly, we identify the topic of inputs to retrieve the corresponding event graph. Depending on tasks, the inputs can be titles for open story generation, or images for the visual storytelling task.
If the input is a piece of text, we directly use the LDA model we trained earlier to identify the topic.
\paragraph{Event Sequence Generation}

Once we identify the topic of input, we walk on the corresponding event graph to generate a sequence of events.
In our experiments, we found that an autoregressive language model tends to produce repetitions, thus resulting in degraded performance. Hence, we propose to use a score-based generation method. 
The algorithm can be seen as a type of beam search, in which the candidate event sequences are ranked by a score function. Starting from a random event $e_1$, we progressively search for the next event $e_t$ in the following candidate set:
\begin{align}
    \{ e_t \, \mid \, & e_t \in \mathrm{Graph}(e_{t-1}), \nonumber \\
    & e_t \notin \mathrm{Exclusive}(e_1,\ldots,e_{t-1}) \},
\end{align}
where $\mathrm{Graph}(\cdot)$ returns a set of possible next events in the graph, $\mathrm{Exclusive}(\cdot)$ returns a set of mutually exclusive events. This filtering step greatly reduces the number of candidate events to consider.

To select the event from the candidate set, we rank all remaining candidate events with the following score function:
\begin{equation}
\begin{aligned}
\mathrm{Score}(e_t) = & \frac{1}{\sum_{i=0}^{t-1}\lambda^i}  \sum_{i=0}^{t-1} \lambda^{i} \log f_{\mathrm{event}}(e_i, e_t)  \\
& + \log f_{\mathrm{input}}(x, e_t)
\label{eq:event}  
\end{aligned}
\end{equation}
where the first term of score function sums the event-event coherence score of candidate event $e_t$ to each partially generated event $e_i$ and gives more weights to recent events. $\lambda$ denotes the decay rate. Then a decayed average applied over the score. The model used in producing the event-event coherence is the same model used in detecting mutually exclusive events.
The second term is an input-event coherence score $f_{\mathrm{input}}(x, e_t)$, which indicates the coherence score between event $e_t$ and the input $x$. 
We propose an input-event coherence model to compute this score. Please refer to Figure~\ref{fig:coherence} for details of parameterization. 
For the task of open story generation, the input-event coherence model is implemented with compositional neural networks, where the input $x$ in Equation~(\ref{eq:event}) is the title. 

As a common practice for beam search, we set a budget of the number of candidates to explore (i.e., beam size). The candidates with the highest scores are maintained in the beam. The final candidate with the highest score is selected as the result of planning. 




\subsection{4.2\quad Story Generation Module}
\label{sec:event-to-story}
The generated event sequence will be sent to a story generation module, which converts the events into a story. 
This story generation module can be any type of model. Recently, large pre-trained language models show great capability of generating knowledgeable and informative sentences. Taking these advantages, the planned events are more likely to be logically connected in the generated story. Therefore, we apply GPT2-small~\cite{gpt2} as our story generation module. During the training, we feed the module with the title words and events. A special token ``$<$EOT$>$'' separates the title and the events and another special token ``$<$SEP$>$'' is placed in every interval of the events. ``$<$$|$endofinput$|$$>$'' token is added at the end of the input. Besides, we also train an RNN-based sequence-to-sequence model that is fed with the same inputs for comparison.

However, as stated in~\citet{plan_write,tan2020progressive}, Exposure bias problem happens when plan-write strategy is applied. 
To mitigate this problem, we alternatively add two kinds of noises into the inputs. 1) Mask 20\% events with a ``[MASK]'' token. 2) Mask all events. The first noise encourages the model to generate sentences referring to all planned events. While, the second noise promotes the model to generate more related stories to the title. The effectiveness of two noises are analyzed in the supplementary material.

%% file: experiment.tex
\section{5\quad Experiment}
\label{section:experiment}

\subsection{5.1\quad Experiment Settings}
We design two experiments to explicitly evaluate event quality and story quality.
Firstly, we calculate the diversity score and conduct human evaluation on the planned events. Secondly, we use the story generation module to transform the events into full stories and conduct human evaluation to evaluate the story quality. 
Moreover, to further verify the correctness of our GraphPlan, we conduct experiments on Story Cloze Test.
The details of model implementations for all experiments can be found in the supplementary material.

\subsection{5.2\quad Dataset}
ROCStories Corpora \cite{rocstory} consists of 98,162 training stories and 1,874 stories for validation and testing. Each story contains a title which we use as the input and a five-sentence story as the target. 
Since titles are annotated only for training data, we split this training set into 8:1:1 for training, validating and testing. 
We applied clustering to the training split (i.e., 8 of 8:1:1) and obtained 500 topics, in which each topic represents one specific domain. Each story is generated from one specific domain in the following experiments. Gold event sequences that are used in planning methods are extracted from the stories in the corpus.

\subsection{5.3\quad Baseline}

\noindent \textbf{S2S} Following~\citet{plan_write}, we use a sequence-to-sequence model \cite{seq2seq}, which straightforwardly generates events given the input titles.

\noindent \textbf{S2S(R)} To build a more competitive baseline, we adopt reward shaping in the sequence-to-sequence model. Like~\cite{tambwekar2018controllable}, we apply policy gradient on 
\begin{equation}
\begin{aligned}
\nabla_{\theta} J(\theta)&=R\left(e_{i}\right) \nabla_{\theta} \log P\left(e_{i}\mid e_{1},\ldots,e_{i-1}; \theta\right)\\
R(v)&=\alpha \times r_{1}(v) \times r_{2}(v)\\
r_{1}(v)&=\log f_{\mathrm{input}}(x, v)\\
r_{2}(v)&=\frac{\sum_{e \in E \land e \neq v} \log f_{\mathrm{event}}(e, v)}{N-1}
\end{aligned}
\end{equation}
where $e$ denotes set of the events in the planning sequence, $E$ denotes the events in the story, $N$ denotes the number of events in the story, $x$ denotes the input title and $\alpha$ denotes the normalization constant across the events in each training sample.
During training, the gradient from $e_{i}$ is multiplied by the Reward $R(e_{i})$ that are proportional to $r_{1}(e_{i})$ and $r_{2}(e_{i})$. In briefly, $r_{1}(e_{i})$ get larger when $e_{i}$ is more related to the input $x$, while $r_{2}(e_{i})$ become larger when $e_{i}$ is more likely to coexist with all events $\{e|e \in E \land e \neq e_{i}\}$.
This method enforces the model to focus on the event that has a high coherence score to the input and events in each training sample.

\noindent \textbf{GR} In this method, we apply random walk on the event graphs while considering mutually exclusive sets. We aims to show the importance of using coherence models by comparing with this method.

\noindent \textbf{GP(Ours)} This is our proposed method that plans events on an event graph with mutually exclusive set and coherence models.

\begin{table}[t!]
  \centering
  \begin{tabular}{l|c|c|c||c}
    \hline
    Diversity&S2S&S2S(R)&GP&GOLD\\    \hline\hline
    Dist-1&10.17\%&11.35\%&\textbf{20.54}\%&24.92\%\\\hline
    Dist-2&56.55\%&58.92\%&\textbf{78.12}\%&87.75\%\\\hline

  \end{tabular}
  \caption{Diversity of planned events. We can see that both sequence-to-sequence models achieve low diversity, while GraphPlan can achieve high diversity.}
  \label{tab:diversity}
\end{table}

\subsection{5.4\quad Event Quality}
We plan the events on randomly selected 1000 test data with different baselines and our proposal. 
We first test the diversity of generated sequences. We calculate Distinct-1 and Distinct-2 scores to show the percentage of unique unigram and bigram events in the whole generated events.
Table~\ref{tab:diversity} shows that the sequence-to-sequence model suffers from producing repeated unigram and bigram events. Graph plan produ more events in the full event set (more unigrams) and produces more combinations between events (more bigrams).

To further evaluate the quality of planned events, we conduct human evaluation. 
Instead of using overly abstract event representation as~\cite{controllable}, we use the verb phrase which is more comprehensive for humans. Thus, we request the annotators to compare the event sequences by two criteria: Relevance and Logicality. 
Table~\ref{tab:open_story_generation_event} shows the human evaluation results. From the results, we can see that our planned events (i.e., verb phrases) are more related to the input title and can be easily transformed into a story.
\begin{table}[!t]
\footnotesize
\centering
\begin{tabular}{c|c|c||c|c||c|c}
    \hline
    Choices(\%) & \multicolumn{2}{c|}{GP vs S2S}& \multicolumn{2}{c|}{GP vs S2S(R)}& \multicolumn{2}{c}{GP vs GR}\\\cline{2-7}
    &GP&  S2S & GP& S2S(R)&GP&GR\\
    \hline
    Revelence& \textbf{47.0}&17.8&\textbf{52.0}&26.0&\textbf{56.3}&12.9\\
    Logical&\textbf{53.5}&14.9 &\textbf{55.2}&26.0 &\textbf{51.5}&17.8\\
    
    \hline
\end{tabular}
\caption{Human evaluation on event planning. Cohen's Kappa coefficient ($\kappa$) for all annotations are in the Moderate agreement (0.4-0.6). Sign tests further show the significant difference. p-values are $<0.01$ for all pairwise comparisons.}
\label{tab:open_story_generation_event}
\end{table}

\begin{table}[t!]
  \centering
  \footnotesize
  \begin{tabular}{l|p{7cm}}
    \hhline{=|=}
    
    Title& Married too fast \\ \hline
    S2S & be$\rightarrow$be$\rightarrow$be$\rightarrow$fall$\rightarrow$marry\\ \hline
    S2S(R) & be$\rightarrow$go(up)$\rightarrow$ask$\rightarrow$say$\rightarrow$marry\\ \hline
    GR& want(do)$\rightarrow$go$\rightarrow$sit$\rightarrow$wonder$\rightarrow$call\\ \hline
    GP & feel$\rightarrow$decide$\rightarrow$begin$\rightarrow$start$\rightarrow$regret\\  \hhline{=|=}
    
    Title& New glasses \\ \hline
    S2S & sit$\rightarrow$be(unhappy)$\rightarrow$have$\rightarrow$go$\rightarrow$find\\ \hline
    S2S(R)& break$\rightarrow$need$\rightarrow$go$\rightarrow$get$\rightarrow$be(glad)\\ \hline
    GR& wake(up)$\rightarrow$be$\rightarrow$(not)care$\rightarrow$take$\rightarrow$ make\\ \hline
    GP& buy$\rightarrow$wear$\rightarrow$break$\rightarrow$shatter$\rightarrow$decide(buy)\\ \hhline{=|=}
    
    Title&The new dress\\ \hline
    S2S &  want$\rightarrow$find$\rightarrow$decide(make)$\rightarrow$buy$\rightarrow$have\\ \hline
    S2S(R)& skip$\rightarrow$go$\rightarrow$have$\rightarrow$let$\rightarrow$be(black)$\rightarrow$make\\ \hline
    GR& celebrate$\rightarrow$wear$\rightarrow$look$\rightarrow$pick(out)$\rightarrow$wear$\rightarrow$make\\ \hline
    GP& want(dress)$\rightarrow$want(change)$\rightarrow$dress$\rightarrow$wear$\rightarrow$feel(beautiful)\\ \hhline{=|=}
    
    Title& Grilled cheese\\ \hline
    S2S & love$\rightarrow$be$\rightarrow$decide$\rightarrow$forget$\rightarrow$end(up)\\ \hline
    S2S(R)& make$\rightarrow$get$\rightarrow$go$\rightarrow$go$\rightarrow$look$\rightarrow$see\\ \hline
    GR& feel(comfortable)$\rightarrow$like$\rightarrow$smile$\rightarrow$decide$\rightarrow$feel(full)\\ \hline
    GP& melt$\rightarrow$put$\rightarrow$decide(roast)$\rightarrow$burn$\rightarrow$taste\\ \hhline{=|=}    

  \end{tabular}
 \caption{Examples of the planned events.}
  \label{tab:plan_example}
 \end{table}

\begin{table*}[!t]
\centering
\footnotesize
\begin{tabular}{c|c|c||c|c||c|c||c|c||c|c}
    \hline
    Choices(\%) & \multicolumn{2}{c|}{GP vs GPT2}& \multicolumn{2}{c|}{GP vs S2S}& \multicolumn{2}{c|}{GP vs S2S(R)}& \multicolumn{2}{c|}{GP vs GR} & \multicolumn{2}{c}{GP vs GP+RNN}\\\cline{2-11}
    &GP& GPT2 & GP& S2S&GP& S2S(R)&GP& GR&GP& GP+RNN\\
    \hline
    Revelence&              33.3&\textbf{41.6}&     \textbf{38.4}&28.3&     \textbf{40.8}&20.3&     \textbf{67.8**}&23.2&        \textbf{37.2*}&23.3\\
    Interestingness&        \textbf{47.9}&41.6&     \textbf{40.1}&30.8&     \textbf{43.2}&34.0&     \textbf{60.7*}&39.3&         \textbf{44.2**}&24.5\\
    Logicality&                \textbf{64.6**}&27.1&   \textbf{42.6**}&19.5&     \textbf{44.6*}&25.0&     \textbf{62.5**}&33.9&     \textbf{37.2}&32.6\\
    Overall&                \textbf{56.3*}&35.4&    \textbf{42.6**}&21.6&     \textbf{42.3*}&24.6&    \textbf{64.3**}&30.4&     \textbf{46.5}&37.2\\
    \hline
\end{tabular}
\caption{Human evaluation on open story generation. (+GPT2) are omitted for all methods expect for GP+RNN. We calculate Cohen's Kappa coefficient ($\kappa$). They are in the Moderate agreement (0.4-0.6) and a few of them are in the Fair agreement (0.2-0.4). We also conduct sign tests to show the significant difference as well. (*) denotes p-value is $<0.05$ and (**) denotes p-value is $<0.01$. The result shows that our method achieves significant improvement on logicality, thereby achieving better overall performance.}
\label{tab:open_story_generation_human}
\end{table*}

Table~\ref{tab:plan_example} shows some of the examples generated by different methods. The results show that sequence-to-sequence models tend to generate repetitive events. Specifically, it tends to output the event that occurs with high frequency in the corpus, such as ``be''(there is sth.) and ``go'' (sb. go somewhere). This is common for a model trained under the framework of maximum likelihood estimation method.
Although reward shaping (S2S(R)) helps a lot, the problem is still not eliminated. Without the limitation of coherence score, GR walks on graphs randomly to produce a sequence. As the graph is relatively not small, achieving a good event sequence is extremely challenging.
We can see that our proposal GP produces more logical and diverse events, which humans can easily tell a story based on these given events.

\subsection{5.5\quad Open Story Generation}
\begin{table}[t!]
  \footnotesize
  \centering
  \begin{tabular}{l|p{7.5cm}}
    \hline
    &
    {\bf Title:} Fire next door
    \\
    \hhline{=|=}
    \multirow{1}{*}[-1.5ex]{\rotatebox{90}{GPT-2}}
    &
    \it{After the fire, the fire was out. The family who lived next door started to panic, so they followed it. [The firehouse across the street was still burning. The fire house was still burning next door.] }(Repetition)
    \\
    \hline
    \multirow{1}{*}[-1.5ex]{\rotatebox{90}{S2S}}
    &
    make $\rightarrow$ think $\rightarrow$ go $\rightarrow$ see $\rightarrow$ get $\rightarrow$ apologize\\ \cline{2-2}
    &
    \it{Lisa's new neighbor \red{made} a big fire outside his house. Lisa \red{thought} she would be sorry. She \red{went} outside to check on him. She \red{saw} that he was not there. Lisa's neighbor could not \red{get} a word with him and \red{apologized}.}
    \\
    \hline
    \multirow{1}{*}[-1.5ex]{\rotatebox{90}{S2S(R)}}
    &
    try $\rightarrow$ tell $\rightarrow$ leave $\rightarrow$ throw $\rightarrow$ take\\ \cline{2-2}
    &
    \it{Sally \red{tried} to fight a burning bush. The homeowner \red{told} her she couldn't do it. She \red{left} a piece of wood in the bush next door. A woman \red{threw} the wood at sally. Sally \red{took} the wood to the fire department.}
    \\
    \hline
    \multirow{1}{*}[-2ex]{\rotatebox{90}{GR}}
    &
    quit $\rightarrow$ be $\rightarrow$ look(buy) $\rightarrow$ find $\rightarrow$ go\\ \cline{2-2}
    &
    \it{I \red{quit} my job when i was 16. The only thing I ever didn't do \red{was} take care of my children. Now I'm \red{looking to buy} a car. I \red{found} a cheap car on ebay. Now i can \red{go} out in the sun every day.}
    \\
    \hline
    \multirow{1}{*}[-1ex]{\rotatebox{90}{GP+RNN}}
    &
    catch(on) $\rightarrow$ come $\rightarrow$ go(off) $\rightarrow$ evacuate $\rightarrow$ extinguish\\ \cline{2-2}
    &
    \it{The fire alarm was \red{caught on}. The firefighters \red{came} to help. The fire alarm \red{went off}. Everyone \red{evacuated}. The firemen \red{extinguished} it before it could go off.}
    \\
    \hline
    \multirow{1}{*}[-1ex]{\rotatebox{90}{GP}}
    &
    catch(on) $\rightarrow$ come $\rightarrow$ go(off) $\rightarrow$ evacuate $\rightarrow$ extinguish\\ \cline{2-2}
    &
    \it{The house next door \red{caught on} fire. The fire department \red{came} to the scene. The fire alarm \red{went off}. The entire neighborhood \red{evacuated}. The fire department \red{extinguished} the fire.}
    \\
    \hline
    \multirow{1}{*}[-2ex]{\rotatebox{90}{Gold}}
    &
    \it{John woke up smelling like something was burning. He went outside. He saw the fire next door. He called the authorities. The firemen came to put out the fire.}
    \\
    \hline
  \end{tabular}
 \caption{Examples of open story generation. The red word represents the events.}
  \label{tab:open_story_generation_example}
 \end{table}

The ultimate goal of event planning is to generate more relative and logical coherent stories. Human evaluation on the event sequence is subjective and tricky since the event is highly abstract. To prove that better event planning improves the story quality, we generate stories using the planned events and conduct human evaluation to assess the Relevance, Logicality, Interestingness and Overall scores. We use story generation module (i.e. GPT2 and RNN) to transform these planned events into the full stories.

We compare the following methods in this experiment. 

\noindent \textbf{GPT2}.
It is a large scale language model that shows great performance in generating stories in recent research. In this method we directly input the title to the GPT2 and generate the full stories.

\noindent \textbf{*+GPT2}.
We associate the aforementioned event planning methods with GPT2 which is used in the story generation module. Thus, we compare with S2S+GPT2, S2S(R)+GPT2, GR+GPT2 and GP+GPT2.

\noindent \textbf{GP+RNN}.
In this method, we use an RNN based sequence-to-sequence model to generate the full story which takes the title and the events as inputs. We compare this method to GP+GPT2 to show the effectiveness of large scale language models in transforming the events into the stories.

We conduct human evaluation on Amazon Mechanical Turk (AMT) over four aspects: {\it Relevance} (whether the story is related to the topic), {\it Interestingness} (whether the story content and style are interesting), {\it Logicality} (whether the story is logical), and {\it Overall} (overall quality). The full details of human evaluation are listed in the supplemantary materials.
We randomly sample 300 titles from the testing set and generate the stories via each method. 
Pairwise comparison is conducted to each criteria and each sample is assigned for two different workers to avoid randomness or personal bias.
Table~\ref{tab:open_story_generation_human} shows that our approach performs better in Logicality, and Overall. Especially, our method greatly outperforms other planning methods in the Logicality measure, which suggests that our planned events are logically sound.
We believe that two factors are the primary reasons of improved logic: 1) each event graph is built from the corpus and thus walking on the graph remains the events' logical relations, and 2) The coherence models filter the candidates and the mutually exclusive set further eliminates the non-logical combinations when planning the events. 
Table~\ref{tab:open_story_generation_example} shows an example of stories generated by these methods. We show both the planned events and the stories. Directly using GPT2 produces repeated sentences. Both equipped with an auto-regressive model for event planning, the events planned by S2S and S2S(R) fail to output satisfactory results and thus results in the low logicality score in the generated sentences. Since there is no restriction on the event selection in GR, the event produced could be irrelevant to the title and even mutually contradictory. By using our proposed method, GP can plan a reasonable set of events and thus generate the most logical story.

\begin{table}[!t]
\begin{center}
\begin{tabular}{l||r|r}
\hline
ACC(\%) & Test v1.0 & Test v1.5
\\ 
\hhline{=||=|=}
DN & 77.60 & 64.45\\
\hline
DN+Origin &  78.87 & 67.64\\
\hline
DN+RandomWalk &  79.36 & 68.09\\
\hline
DN+GraphPlan &  \textbf{80.15} & \textbf{69.45}\\
\hline
\end{tabular}
\end{center}
\caption{Results on story cloze test. DN, denote the DiffNet. From the results, events planned by our event graphs and mutually exclusive sets have positive effects on this task.}
\label{tab:sct}
\end{table}

\subsection{5.6\quad Story Cloze Test}
To better validate the effectiveness of our event graphs and mutual exclusive relation between events, we conduct Story Cloze Test. This task aims to select the right ending sentence from two candidates.
We incorporate the event feature generated by different methods into the story cloze test.
The accuracy of Story Cloze Test would reflect the quality of the event features. 
The event feature is learned by a mask language model (MLM) (i.e., BERT model with fewer parameters)~\cite{bert}: if the training event sequence is more logical and reasonable, the learned feature by MLM would better fit the story cloze test. To prove that our event graph and mutually exclusive relation can help us to generate reasonable event sequences, we compare the features generated by MLM model with different training data: (1) \textbf{Origin}: event sequences extracted from ROCStories Corpora. (2) \textbf{RandomWalk}: Randomly walk on the event graphs and sample training data. (3) \textbf{GraphPlan}: Use our planning method to generate training data. Note that the input-event coherence score is excluded in the score function due to no input being given.
We then use the event feature and adopt the state-of-the-art model DiffNet \cite{cui2019discriminative} for the story cloze test.
For fair comparison, RandomWalk and GraphPlan sample the same amount of event chains in the dataset during training. Further details of the model could be found in the supplementary material.

Results of Story Cloze Test are presented in Table~\ref{tab:sct}. It shows that RandomWalk and GraphPlan achieves better scores in both SCTv1.0 and v1.5, which prove that our event graphs and mutually exclusive events have positive effects on event planning.


\subsection{5.7\quad Other Story Generation Task}
To further verify the effectiveness of our proposed method, we also conduct experiments on Visual Storytelling tasks. Due to the page limitation, we put it into the supplementary material. In the results, GraphPlan also shows improvement in terms of logicality.

%% file: discussion.tex
\section{6\quad Discussion}

\begin{table}[!t]
  \centering
  \begin{tabular}{p{8.0cm}}
 
    \includegraphics[width=7cm, height=1.6cm]{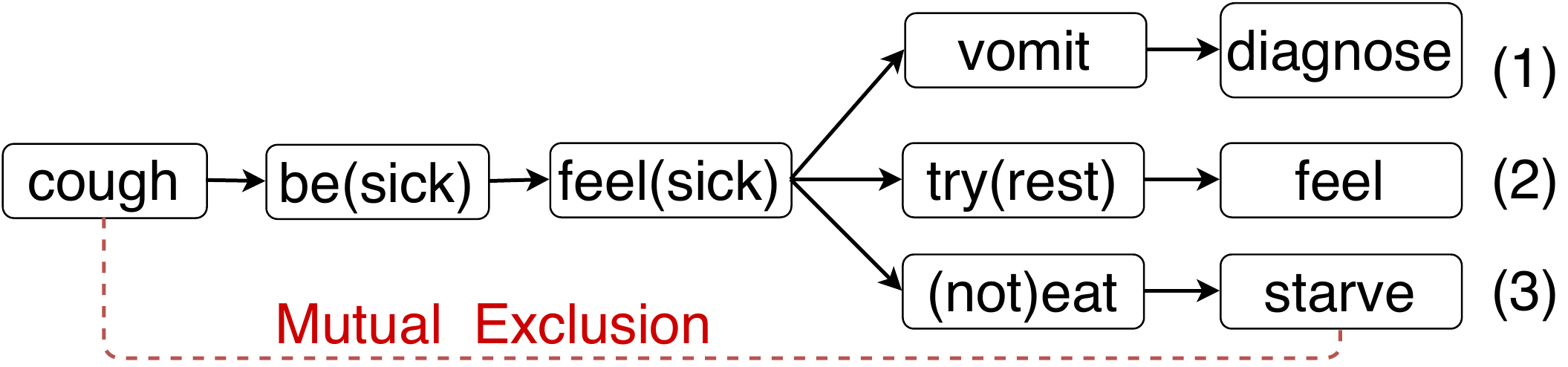}
    \\ \hline
    (1) \it{The man was \red{coughing} a lot. He \red{was sick}. He \red{felt sick} for days. He \red{vomited} on the couch. He \red{was} later \red{diagnosed} with the flu.}
    \\
    \hline
    (2) \it{The man was \red{coughing} a lot. He \red{was sick}. He \red{felt sick}. He \red{tried} to \red{rest} for an hour. The man \red{felt} better !}
    \\
    \hline
    (3) \it{The man was \red{coughing} a lot. He \red{was sick}. He \red{felt sick}. He \red{couldn't eat} anything. \underline{\underline{He \red{starved} himself.}}}
    \\
    \hline
  \end{tabular}
  \caption{Example of controllable generation. (1) and (2) extends different events after ``feel sick'' to achieve different endings and (3) shows logical inconsistency when generating with two mutually exclusive events}
  \label{tab:control}
 \end{table}

\paragraph{Controllable Generation}
As mentioned before, the stories can be easily controlled by modifying the events. Table~\ref{tab:control} shows an example. Selecting different upcoming events for ``feel(sick)'' would change the following storyline.

\paragraph{Why GraphPlan Works?}
Our experiment shows that event graphs can produce more logical stories than planning with language models. Here we give an empirical explanation.
Sequence-to-sequence models usually fail to capture long term relation and order information in the event sequence. The decoder is not guaranteed to take into account all previous events during decoding. 
At this point, our approach applying event-event coherence scores enforces the model to consider long term relations during planning. In addition, the order of events is captured from the gold cases that can be guaranteed in our event graphs. 
Moreover, mutually exclusive sets help us to decide whether two events can co-occur in one sequence no matter how distant two events are.
Table~\ref{tab:control} gives an example. ``cough'' and ``starve'' are considered as mutually exclusive events in our event graph. If we generate a story based on this event chain, the last sentence ``he starved himself'' seems not reasonable in this case.

The findings in this work also opens up new research questions: 1) Better event definition 2) Exposure Bias issue in the story generation module 3) Better topic modelling methods. These are left for our future work.

%% file: conclusion.tex
\section{7\quad Conclusion}


In this paper, we show that a graph-based event planning approach can indeed produce more natural event sequences compared to conventional language models. We propose to walk on automatically learned event graphs by performing beam search with a score function dedicated for event planning. Then the story is generated follows the planned events.

We evaluate our approach on event planning and open story generation with large scale human judgements. Results show that our proposed approach clearly outperforms the non-planning baseline and the sequence-to-sequence model based planning models. In human evaluation, the events and the stories generated by our proposal are believed to be more logical and coherent. An additional experiment on story cloze test further proves the advantages of event graphs and mutually exclusive sets.


